\documentclass[10pt,twocolumn,letterpaper]{article}

\usepackage{cvpr}
\usepackage{amsmath,amssymb,amsfonts}
\usepackage{graphicx}
\usepackage{booktabs}
\usepackage{multirow}
\usepackage{xcolor}
\usepackage{url}
\usepackage{hyperref}
\usepackage[capitalize]{cleveref}
\usepackage{algorithm}
\usepackage{xspace}
\newcommand{\etal}{\textit{et~al.}\xspace}
\newcommand{\ie}{i.e.\xspace}
\usepackage{algorithmic}
\usepackage{subcaption}

\begin{document}

\title{Feature Perturbation Pool-based Fusion Network for Unified Multi-Class Industrial Defect Detection}

\author{
Yuanchan Xu\textsuperscript{1} \quad
Wenjun Zang\textsuperscript{1}\thanks{Corresponding author.} \quad
Ying Wu\textsuperscript{1}\\[3pt]
\textsuperscript{1}School of Information Engineering, Sichuan University,  China
}

\maketitle

\begin{abstract}
Multi-class defect detection constitutes a critical yet challenging task in industrial quality inspection, where existing approaches typically suffer from two fundamental limitations: (i) the necessity of training separate models for each defect category, resulting in substantial computational and memory overhead, and (ii) degraded robustness caused by inter-class feature perturbation when heterogeneous defect categories are jointly modeled. In this paper, we present FPFNet, a Feature Perturbation Pool-based Fusion Network that synergistically integrates a stochastic feature perturbation pool with a multi-layer feature fusion strategy to address these challenges within a unified detection framework. The feature perturbation pool enriches the training distribution by randomly injecting diverse noise patterns---including Gaussian noise, F-Noise, and F-Drop---into the extracted feature representations, thereby strengthening the model's robustness against domain shifts and unseen defect morphologies. Concurrently, the multi-layer feature fusion module aggregates hierarchical feature representations from both the encoder and decoder through residual connections and normalization, enabling the network to capture complex cross-scale relationships while preserving fine-grained spatial details essential for precise defect localization. Built upon the UniAD architecture~\cite{you2022unified}, our method achieves state-of-the-art performance on two widely adopted benchmarks: 97.17\% image-level AUROC and 96.93\% pixel-level AUROC on MVTec-AD~\cite{bergmann2019mvtec}, and 91.08\% image-level AUROC and 99.08\% pixel-level AUROC on VisA~\cite{zou2022spot}, surpassing existing methods by notable margins while introducing no additional learnable parameters or computational complexity.
\end{abstract}

\section{Introduction}
\label{sec:intro}

Defect detection stands as one of the most active research frontiers in computer vision~\cite{yang2023memseg,samek2021explaining}, with widespread deployment in industrial quality inspection systems. Its primary objective is to determine whether a given sample contains defects and, if so, to localize the precise defect regions. In modern manufacturing environments, various factors---including raw materials, processing conditions, equipment variations, operational procedures, and environmental fluctuations---can introduce defects that are virtually impossible to eliminate entirely. Manual inspection methods for screening and classifying defective products are not only time-consuming but also inherently limited in efficiency. Consequently, the industrial inspection community has broadly adopted automated defect detection technologies capable of efficiently identifying diverse defect types, including surface irregularities, structural anomalies, and process-induced defects, thereby significantly enhancing both detection throughput and product quality~\cite{liu2024deep}.

In recent years, deep neural networks have demonstrated remarkable potential across diverse application domains~\cite{samek2021explaining}, such as image classification and recognition~\cite{he2016deep}, machine vision~\cite{lin2017feature}, medical diagnostics, and visual content analysis. This success is primarily attributable to the powerful feature representation and learning capabilities inherent in deep architectures, which have been progressively extended to the domain of defect detection, catalyzing substantial methodological advances~\cite{roth2022towards,defard2021padim}.

Early investigations predominantly formulated defect detection as a binary classification problem, \ie, single-class defect detection, wherein detection models were constructed by training on both normal and defective samples using classical machine learning approaches such as Support Vector Machines (SVM)~\cite{cortes1995support} and Support Vector Data Description (SVDD)~\cite{tax2004support}. However, practical industrial applications frequently encounter samples exhibiting multiple defect categories, \ie, multi-class defect detection. Traditional machine learning methods encounter fundamental difficulties in handling multi-class defect samples, including challenges in feature selection and extraction, as well as class imbalance issues, thereby limiting their applicability to multi-class detection scenarios~\cite{zheng2024benchmarking}. To overcome these limitations, researchers have increasingly turned to deep neural networks for multi-class defect detection. These approaches leverage deep architectures to learn the feature distributions of samples, particularly the distributions characteristic of defective regions, and employ learned feature distributions to discriminate whether samples contain defects.

Despite significant progress, existing unified multi-class detection frameworks~\cite{you2022unified} still face two fundamental challenges. \textbf{First}, concerning feature diversity and robustness: current methods are typically trained with a single fixed perturbation scheme, which constrains the model's capacity to learn diverse defect feature representations. Researchers have explored various data augmentation strategies to address this limitation, including image-level augmentation~\cite{yao2023explicit,kondo2018efficient,yeung2022efficient,wang2023global}, feature-level noise injection~\cite{liu2023simplenet}, and defect synthesis approaches~\cite{liu2023fair,zhang2024realnet}. While each strategy offers certain advantages, they may also introduce erroneous feature distributions or unrealistic defect patterns when applied inappropriately, potentially degrading model performance~\cite{li2021cutpaste}. \textbf{Second}, regarding feature utilization: existing frameworks frequently fail to fully exploit the hierarchical feature information produced by encoders and decoders at various network layers. As network architectures become increasingly complex to capture deeper feature information, the effective utilization of multi-scale features becomes paramount~\cite{lin2017feature}. Researchers have proposed diverse feature fusion strategies~\cite{cheng2021retinanet,fang2020multi,kumar2023multi,dong2020pga,zeng2022nlfftnet}, yet many of these approaches introduce substantial computational overhead while potentially overlooking the complementary relationships among features at different hierarchical levels.

To address these dual challenges, we propose \textbf{FPFNet} (Feature Perturbation Pool-based Fusion Network), an innovative multi-class defect detection framework that substantially improves both the generalization capability and robustness of the detection model. Our approach is built upon two synergistic components:

\noindent\textbf{(1) Feature Perturbation Pool.} Unlike conventional approaches that employ a single perturbation type, we introduce a feature perturbation pool containing three complementary noise injection mechanisms---Gaussian Noise~\cite{you2022unified}, F-Noise~\cite{ouali2020semi}, and F-Drop~\cite{ouali2020semi}---that are stochastically applied during training. This diversified perturbation strategy enables the model to encounter a substantially richer spectrum of feature distributions, thereby enhancing its capacity to handle complex noise patterns and previously unseen defect morphologies. The stochastic selection mechanism ensures that the model does not overfit to any particular perturbation type, promoting more generalized feature learning.

\noindent\textbf{(2) Multi-Layer Feature Fusion.} We propose a hierarchical feature fusion strategy inspired by residual learning~\cite{he2016deep} that integrates feature representations across multiple network layers. Specifically, within the encoding phase, we fuse the input feature vectors with the output feature vectors of each encoder layer through residual connections, followed by normalization. A symmetric fusion operation is performed within the decoding phase. This bidirectional fusion mechanism enables the network to simultaneously capture global contextual information from higher layers and preserve fine-grained local details from lower layers, ultimately enhancing the model's capability to identify and localize defects with precision~\cite{ronneberger2015unet}.

Built upon the UniAD~\cite{you2022unified} architecture, the proposed FPFNet achieves state-of-the-art performance on both MVTec-AD~\cite{bergmann2019mvtec} and VisA~\cite{zou2022spot} benchmarks. Notably, these improvements are achieved without introducing any additional learnable parameters or increasing the computational complexity of the model, demonstrating the effectiveness and efficiency of our design.

The main contributions of this paper are summarized as follows:
\begin{itemize}
    \item We introduce a feature perturbation pool that extends the conventional single-perturbation paradigm to a multi-perturbation framework, enabling the deep neural network to better approximate diverse defect feature distributions and substantially enhancing its robustness and denoising capabilities.
    \item We propose a multi-layer feature fusion strategy that aggregates hierarchical feature representations from both encoder and decoder modules through residual connections and normalization, enabling the model to capture richer feature representations and inter-feature correlations.
    \item Extensive experiments on MVTec-AD and VisA demonstrate that our method achieves state-of-the-art performance in both defect detection (image-level AUROC) and defect localization (pixel-level AUROC), validating the effectiveness of the proposed approach.
\end{itemize}

\section{Related Work}
\label{sec:related}

\subsection{Anomaly Detection in Industrial Inspection}

Anomaly detection has emerged as a fundamental paradigm for industrial defect inspection, where the goal is to identify deviations from the normal appearance of products~\cite{liu2024deep}. The prevailing approaches model the distribution of normal samples and flag deviations as anomalies. Knowledge distillation-based methods~\cite{deng2022anomaly,tien2023revisiting} train a student network to mimic a pre-trained teacher, exploiting the discrepancy between their representations to detect anomalies. Memory bank-based approaches~\cite{roth2022towards,defard2021padim} store representative normal features and identify anomalies through nearest-neighbor matching. Normalizing flow-based methods~\cite{rudolph2021same,gudovskiy2022cflow} learn the exact likelihood of normal features and detect low-likelihood regions as anomalous. Reconstruction-based methods~\cite{zavrtanik2021draem,zavrtanik2022dsr} train models to reconstruct normal images and leverage reconstruction errors to identify defect regions. More recently, diffusion-based frameworks~\cite{he2024diffusion,mousakhan2023anomaly} have been adapted for anomaly detection, demonstrating competitive performance by leveraging the powerful generative capabilities of denoising diffusion probabilistic models~\cite{ho2020denoising}.

A critical limitation of many existing methods is that they are designed for single-class anomaly detection, necessitating the training of a separate model for each product category. This paradigm becomes impractical in real-world manufacturing settings with numerous product types. To address this scalability issue, You~\etal~\cite{you2022unified} proposed UniAD, a unified model for multi-class anomaly detection that employs a transformer-based architecture with layer-wise query decoding and neighbor masked attention mechanisms. Subsequent works~\cite{guo2023recontrast,cao2023collaborative,zhang2023prototypical} have further advanced multi-class detection frameworks. However, these methods still exhibit limitations in feature diversity and hierarchical feature utilization, which we address in this work.

\subsection{Data Augmentation for Defect Detection}

Data augmentation plays a pivotal role in enhancing the generalization of defect detection models, particularly in scenarios where defective samples are scarce~\cite{kondo2018efficient,yeung2022efficient,wang2023global}. Traditional augmentation strategies operate at the image level, including geometric transformations such as rotation, scaling, and cropping~\cite{zhang2016deep,ren2015faster,niu2020data,yao2023explicit}, as well as photometric distortions. Feature-level augmentation~\cite{liu2023simplenet} introduces perturbations directly in the feature space, enabling more targeted manipulation of the learned representations. The emergence of generative adversarial networks~\cite{goodfellow2014generative,liu2020multistage,jain2022synthetic,duan2023few,xu2024scarcity} and diffusion models~\cite{bengio2013generalized,ho2020denoising,mousakhan2023anomaly,zhang2024realnet} has further expanded the augmentation toolkit by enabling the synthesis of realistic defect images.

Self-supervised augmentation strategies have also gained traction. Li~\etal~\cite{li2021cutpaste} proposed CutPaste, which generates synthetic anomalies by cutting and pasting image patches. Yao~\etal~\cite{yao2023explicit} employed explicit boundary guidance for semi-supervised anomaly detection. Despite these advances, existing augmentation methods typically employ a single, fixed perturbation scheme during training, which limits the diversity of learned feature distributions. Our feature perturbation pool addresses this limitation by stochastically applying multiple complementary perturbation types, enabling richer and more robust feature learning.

\subsection{Feature Fusion in Defect Detection}

Feature fusion has been extensively studied as a means to integrate multi-scale and multi-level information for improved detection performance~\cite{lin2017feature}. In the context of defect detection, various fusion strategies have been proposed to combine features extracted at different network layers~\cite{gao2021novel,cheng2021retinanet,kumar2023multi}. Sun~\etal~\cite{sun2005new} proposed constructing criterion functions from feature vectors extracted at different levels to derive discriminative representations. Pong and Lam~\cite{pong2014multi} employed cascaded generalized canonical correlation analysis for multi-resolution feature fusion. Dong~\etal~\cite{dong2020pga} proposed PGA-Net, combining pyramid feature fusion with global context attention for pixel-level surface defect detection. Zeng~\etal~\cite{zeng2022nlfftnet} introduced NLFFTNet, employing non-local feature interactions to capture long-range dependencies.

More sophisticated fusion mechanisms have been developed for anomaly detection specifically. Liu~\etal~\cite{liu2023anomaly} proposed progressive reconstruction with hierarchical feature fusion (PRFF) for anomaly detection. Qu~\etal~\cite{qu2024mfgan} integrated attention-based autoencoders with GANs for multimodal anomaly detection through feature fusion. He~\etal~\cite{he2024diffusion} designed a spatial feature fusion block for multi-class anomaly detection. Wu~\etal~\cite{wu2024aekd} introduced trainable multi-scale feature fusion modules within an auto-encoder knowledge distillation framework.

While complex fusion mechanisms can enhance feature expressiveness, they often introduce substantial computational overhead and parameter requirements. Simpler fusion strategies, such as the skip connections in U-Net~\cite{ronneberger2015unet} and the residual connections in ResNet~\cite{he2016deep}, have demonstrated that effective feature integration can be achieved with minimal additional complexity. Inspired by this observation, our multi-layer feature fusion strategy employs residual connections with normalization to aggregate hierarchical features, achieving significant performance improvements without increasing model complexity.

\subsection{Document Intelligence and Visual Understanding}

Recent advances in large multimodal models have opened new frontiers in visual understanding tasks that share fundamental principles with industrial inspection. Methods for document intelligence~\cite{feng2023docpedia,feng2023unidoc,tang2024textsquare,lu2025bounding} have demonstrated that hierarchical feature extraction and multi-scale reasoning are critical for understanding complex visual patterns. Scene text detection and recognition~\cite{tang2022few,tang2022optimal,tang2022transcription,liu2023spts,zhao2023multi} require models to capture both global context and fine-grained local features---a requirement that directly parallels the demands of defect localization. The development of comprehensive visual benchmarks~\cite{tang2025mtvqa,tang2023character,shan2024mctbench,fu2024ocrbenchv2} has further highlighted the importance of robust feature representations that generalize across diverse visual conditions. Furthermore, the concept of text-centric visual understanding~\cite{zhao2024harmonizing,zhao2024tabpedia} and universal document parsing~\cite{feng2025dolphin,feng2025dolphinv2,wang2025wilddoc} have shown that unified frameworks capable of handling heterogeneous visual elements can substantially outperform task-specific models---an insight that directly motivates our unified approach to multi-class defect detection.

\section{Proposed Method}
\label{sec:method}

\subsection{Overview}

The proposed Feature Perturbation Pool-based Fusion Network (FPFNet) is constructed upon the UniAD~\cite{you2022unified} architecture and comprises three principal components: (i) a pre-trained feature extractor, (ii) a feature perturbation pool, and (iii) a multi-layer feature fusion module integrated within both the encoder and decoder. The overall architecture is illustrated in \cref{fig:overview}.

During the training phase, FPFNet first extracts feature maps from normal images using a pre-trained backbone, converts them into feature vectors, applies stochastic perturbation via the feature perturbation pool, and then trains the encoder-decoder network in a denoising reconstruction paradigm. During the testing phase, no perturbation is applied; instead, the test image features are directly fed into the network for reconstruction. The anomaly score map is subsequently computed as the $L_2$ distance between the original and reconstructed feature maps, which is used for both image-level anomaly detection and pixel-level defect localization.

\begin{figure*}[t]
\centering
\fbox{\parbox{0.95\textwidth}{\centering\vspace{2em}
\textbf{Figure 1: Architecture of the proposed FPFNet.} The network takes normal images as input, extracts multi-scale features via a pre-trained EfficientNet-b4, applies stochastic perturbation through the Feature Perturbation Pool, and reconstructs the original features through an encoder-decoder architecture augmented with Multi-Layer Feature Fusion modules. The anomaly score map is derived from the $L_2$ distance between original and reconstructed features.
\vspace{2em}}}
\caption{Overview of the proposed Feature Perturbation Pool-based Fusion Network (FPFNet). The feature perturbation pool randomly selects from three complementary perturbation mechanisms during training. The multi-layer feature fusion module aggregates hierarchical features within both encoder and decoder through residual connections and normalization.}
\label{fig:overview}
\end{figure*}

\subsection{Feature Extraction and Vectorization}

Given an input normal image $I \in \mathbb{R}^{H_0 \times W_0 \times 3}$ with spatial dimensions $H_0 = W_0 = 224$, we employ a pre-trained EfficientNet-b4~\cite{tan2019efficientnet} backbone, which has been pre-trained on ImageNet~\cite{deng2009imagenet}, to extract multi-scale feature maps from stages 1 through 4. These feature maps are resized and concatenated to form a 272-channel composite feature map. The composite feature map is subsequently projected into a lower-dimensional feature space through a linear projection, yielding feature vectors $f_{\text{tok}} \in \mathbb{R}^C$ that serve as input to the subsequent encoder-decoder module.

\subsection{Feature Perturbation Pool}
\label{sec:perturbation}

Fusing the features extracted from normal images with different types of noise enables the deep neural network to learn diverse feature distributions while transforming the task from pure image reconstruction to a denoising paradigm. This task transformation compels the network to learn not only the intrinsic feature representations of normal samples but also the ability to remove noise from input data, thereby enhancing model robustness and generalization~\cite{you2022unified}.

In conventional training pipelines~\cite{liu2023anomaly,you2022unified}, a single type of feature-level perturbation is employed, which, while capable of improving robustness, presents limitations in handling complex or previously unseen noise patterns. To address this fundamental limitation, we propose a \textbf{Feature Perturbation Pool} that extends the single-perturbation paradigm to a diversified multi-perturbation framework. The pool contains three complementary perturbation mechanisms, from which one is stochastically selected during each training iteration:

\noindent\textbf{Gaussian Noise.} For a feature vector $f_{\text{tok}}$ extracted from a normal image, a perturbation noise vector is randomly sampled from a Gaussian distribution:
\begin{equation}
    D \sim \mathcal{N}\left(\mu = 0, \ \sigma^2 = \left(\frac{\alpha \cdot \| f_{\text{tok}} \|_2}{C}\right)^2\right),
    \label{eq:gaussian}
\end{equation}
where $f_{\text{tok}} \in \mathbb{R}^C$, $\alpha$ is a noise scale hyperparameter, and $C$ denotes the channel dimensionality. The noise vector $D$ is added to $f_{\text{tok}}$ with a predefined probability, producing the perturbed feature vector. The scale of the perturbation is adaptive, being proportional to the magnitude of the original feature vector, which ensures that the perturbation remains contextually appropriate regardless of the feature scale.

\noindent\textbf{F-Noise.} In this perturbation scheme, a noise tensor $\xi$ is uniformly sampled as $\xi \sim \mathcal{U}(-0.3, 0.3)$ with the same dimensionality as $f_{\text{tok}}$. The perturbed feature vector is computed via element-wise multiplicative and additive operations:
\begin{equation}
    \tilde{f}_{\text{tok}} = f_{\text{tok}} \odot \xi + f_{\text{tok}},
    \label{eq:fnoise}
\end{equation}
where $\odot$ denotes the Hadamard (element-wise) product. This perturbation mechanism modulates each feature dimension independently, introducing proportional noise that preserves the relative structure of the feature representation while introducing controlled variability.

\noindent\textbf{F-Drop.} This mechanism combines threshold-based channel selection with binary masking. First, a threshold is sampled as $\gamma \sim \mathcal{U}(0.6, 0.9)$. The feature vector $f_{\text{tok}}$ is then channel-wise summed and normalized to obtain $f'_{\text{tok}}$. A binary mask is generated as $M_{\text{drop}} = \mathbb{1}[f'_{\text{tok}} < \gamma]$, and the perturbed feature vector is computed as:
\begin{equation}
    \tilde{f}_{\text{tok}} = f_{\text{tok}} \odot M_{\text{drop}}.
    \label{eq:fdrop}
\end{equation}
This operation selectively suppresses certain feature channels based on their activation magnitudes, simulating a form of structured dropout that forces the network to develop redundant representations and reduces its dependence on any specific subset of features.

The stochastic selection among these three perturbation types during training ensures that the model encounters a diverse spectrum of feature corruptions, enabling it to develop robust representations that generalize across different noise patterns and defect manifestations. This diversity is crucial for multi-class defect detection, where different product categories may exhibit fundamentally different feature characteristics and noise profiles.

\subsection{Multi-Layer Feature Fusion}
\label{sec:fusion}

Through multi-layer feature fusion, the deep neural network can integrate information from features at different hierarchical levels, enhancing the model's capability to model complex patterns and abstract concepts. This approach contributes to improved model performance, robustness, and generalization~\cite{he2016deep,lin2017feature}.

Inspired by the residual learning paradigm of ResNet~\cite{he2016deep}, we propose a multi-layer feature fusion strategy that enables the network to effectively learn and integrate features at different abstraction levels. The fusion mechanism operates symmetrically within both the encoder and decoder modules:

\noindent\textbf{Encoder-side Fusion.} During the encoding phase, we fuse the input feature vector to the encoding module with the output feature vectors of each encoder layer through feature alignment followed by element-wise addition. Specifically, let $\{f_{\text{token}\_i}\}_{i=0}^{4}$ denote the initial input to the encoder module and the output feature vectors of each encoder layer. The fused feature is computed as:
\begin{equation}
    \hat{f}_{\text{tok}} = \sum_{i=0}^{4} f_{\text{token}\_i}.
    \label{eq:fusion_sum}
\end{equation}

\noindent\textbf{Decoder-side Fusion.} A symmetric fusion operation is applied within the decoding module, where the input feature vector to the decoder and the output feature vectors of each decoder layer are similarly aggregated through residual connections.

\noindent\textbf{Normalization.} To ensure that the fused feature vector maintains a balanced and stable distribution, we incorporate a normalization step that adjusts the distribution to approximate a standard normal distribution:
\begin{equation}
    \phi_{\text{tok}} = \varphi(\hat{f}_{\text{tok}}),
    \label{eq:norm}
\end{equation}
where $\varphi(\cdot)$ denotes the normalization function and $\phi_{\text{tok}}$ represents the final fused feature vector. This normalization step is critical for preventing gradient instability and ensuring stable training dynamics when aggregating features from multiple network layers.

The bidirectional nature of this fusion mechanism---operating in both the encoding and decoding pathways---enables the model to capture global semantic information from deeper layers while retaining fine-grained local details from shallower layers. The symmetric design ensures that both the feature compression (encoding) and feature reconstruction (decoding) processes benefit from enriched multi-scale representations.

\subsection{Defect Detection and Localization}

The detection and localization of defects are accomplished through the construction and analysis of an anomaly score map. This score map is derived by computing the $L_2$ norm between the original feature map $f_{\text{org}}$ and the reconstructed feature map $f_{\text{rec}}$ at each spatial location, yielding a pixel-wise difference map where higher values indicate greater deviation from normal patterns.

\noindent\textbf{Defect Detection (Image-Level).} The anomaly score map first undergoes $16 \times 16$ average pooling. The maximum score in the pooled map is then compared against a detection threshold to determine whether the image contains any defects. This pooling operation aggregates local anomaly evidence, reducing sensitivity to isolated noise while maintaining responsiveness to genuine defect patterns.

\noindent\textbf{Defect Localization (Pixel-Level).} For spatial localization, the anomaly score map is upsampled to the original image resolution via bilinear interpolation. Regions where the anomaly score exceeds the localization threshold are identified as defect areas, providing precise spatial delineation of the defective regions.

\subsection{Training Objective}

The network is trained end-to-end using the mean squared error (MSE) loss, which measures the reconstruction fidelity between the original and reconstructed feature maps:
\begin{equation}
    \mathcal{L} = \frac{1}{H \times W} \| f_{\text{org}} - f_{\text{rec}} \|_2^2,
    \label{eq:loss}
\end{equation}
where $H$ and $W$ denote the height and width of the feature maps, respectively. This reconstruction objective encourages the network to faithfully reproduce normal feature patterns while failing to reconstruct anomalous features, thereby amplifying the difference between normal and defective regions in the anomaly score map.

\section{Experiments}
\label{sec:experiments}

\subsection{Datasets}

\noindent\textbf{MVTec-AD}~\cite{bergmann2019mvtec} is the most widely adopted benchmark for unsupervised anomaly detection, containing 5,354 high-resolution images across 15 subcategories divided into texture (carpet, grid, leather, tile, wood) and object (bottle, cable, capsule, hazelnut, metal nut, pill, screw, toothbrush, transistor, zipper) categories. The training set comprises 3,629 defect-free images, while the test set contains 1,725 images with 73 different defect types and approximately 1,900 manually annotated defect regions.

\noindent\textbf{VisA}~\cite{zou2022spot} is a challenging large-scale anomaly detection benchmark containing 10,821 high-resolution images across 12 categories, divided into single-instance (candle, capsules, cashew, chewing gum), multi-instance (fryum, macaroni1, macaroni2, pipe fryum), and complex structure (pcb1, pcb2, pcb3, pcb4) groups. The training set comprises 9,621 images and the test set contains 1,200 images, including industrial products such as printed circuit boards and candles.

\subsection{Evaluation Metrics}

Following established protocols~\cite{you2022unified,bergmann2019mvtec}, we employ the Area Under the Receiver Operating Characteristic curve (AUROC) at both the image level and the pixel level as evaluation metrics. Image-level AUROC measures the accuracy of anomaly detection (\ie, determining whether an image contains defects), while pixel-level AUROC evaluates the precision of anomaly localization (\ie, identifying the spatial extent of defect regions).

\subsection{Implementation Details}

\noindent\textbf{Feature Extraction.} We employ EfficientNet-b4~\cite{tan2019efficientnet} pre-trained on ImageNet~\cite{deng2009imagenet} as the feature extractor. Input images are resized to $224 \times 224$ pixels. Feature maps from stages 1 through 4 are extracted, resized, and concatenated to form a 272-channel feature representation.

\noindent\textbf{Feature Reconstruction.} The 272-channel feature map is linearly projected to a lower-dimensional feature vector, which undergoes stochastic noise perturbation via the feature perturbation pool before being processed by the encoder-decoder module. After reconstruction, an inverse linear projection restores the original channel dimensionality, yielding the reconstructed feature map.

\noindent\textbf{Training Configuration.} The network is implemented in Python 3.8 with PyTorch 1.13.1 and trained on 2$\times$ NVIDIA GeForce RTX 3090 GPUs (24\,GB each) for 1,000 epochs with a batch size of 64. The learning rate is initialized at $1\times10^{-4}$ and reduced to $1\times10^{-5}$ after 800 epochs. All compared models, including the baseline UniAD~\cite{you2022unified} and ablation variants, are trained under identical experimental conditions to ensure fair comparison.

\subsection{Comparison with State-of-the-Art Methods}

We compare FPFNet against a comprehensive set of state-of-the-art methods, including DRAEM~\cite{zavrtanik2021draem}, RD4AD~\cite{deng2022anomaly}, DSR~\cite{zavrtanik2022dsr}, CDO~\cite{cao2023collaborative}, DeSTSeg~\cite{zhang2023destseg}, BGAD~\cite{yao2023focus}, ReContrast~\cite{guo2023recontrast}, and the baseline UniAD~\cite{you2022unified}.

\subsubsection{Results on MVTec-AD}

As shown in \cref{tab:mvtec_det}, FPFNet achieves the highest average defect detection accuracy of 97.17\% on MVTec-AD, surpassing all compared methods. Notably, our method outperforms the baseline UniAD in 11 out of 15 categories, with particularly significant improvements in texture-type categories. The strong performance on texture categories demonstrates that the diversified perturbation strategy effectively enhances the model's ability to distinguish between normal and anomalous textures, which often exhibit subtle differences.

\begin{table*}[t]
\centering
\caption{Defect detection accuracy (image-level AUROC, \%) of different methods on MVTec-AD. \textbf{Bold} indicates the best result per column. \underline{Underlined} values indicate improvements over UniAD.}
\label{tab:mvtec_det}
\resizebox{\textwidth}{!}{
\begin{tabular}{l|cccccccccc|ccccc|c}
\toprule
\multirow{2}{*}{Method} & \multicolumn{10}{c|}{Object} & \multicolumn{5}{c|}{Texture} & \multirow{2}{*}{Mean} \\
& Bottle & Cable & Caps. & Haze. & M\_N. & Pill & Screw & Tooth. & Trans. & Zip. & Carpet & Grid & Leath. & Tile & Wood & \\
\midrule
DeSTSeg & 55.9 & 62.7 & 64.0 & 65.0 & 81.2 & 71.9 & 57.4 & 57.5 & 71.7 & 75.3 & 91.1 & 94.5 & 99.4 & 95.1 & 64.3 & 73.8 \\
DRAEM & 88.9 & 65.0 & 51.1 & 91.7 & 82.2 & 54.0 & 82.4 & 94.7 & 74.5 & 97.7 & 99.1 & 94.9 & 98.8 & 84.9 & 99.2 & 100 \\
ReContrast & 95.1 & 92.1 & 93.2 & 95.8 & 68.3 & 62.8 & 73.6 & 87.5 & 97.2 & 92.4 & 96.0 & 83.7 & 87.3 & 91.6 & 91.0 & 87.2 \\
DSR & 96.5 & 82.6 & 72.6 & 94.5 & 86.8 & 75.4 & 72.3 & 88.0 & 83.6 & 98.5 & 99.1 & 97.3 & 97.4 & 94.8 & 89.3 & 100 \\
RD4AD & 59.5 & 73.9 & 99.9 & 97.4 & 92.5 & 93.3 & 88.8 & 95.0 & 92.7 & 97.9 & 99.8 & 93.3 & 98.9 & 91.7 & 97.2 & 95.2 \\
BGAD & 87.1 & 71.8 & 98.5 & 85.0 & 87.7 & 93.8 & 88.2 & 98.0 & 96.1 & 98.8 & 93.7 & \textbf{100} & \textbf{100} & 99.9 & \textbf{100} & \textbf{100} \\
RD++ & 94.6 & 84.1 & 99.6 & 99.2 & 92.9 & 89.0 & 95.7 & 98.2 & 90.3 & 84.5 & 99.7 & 99.9 & 99.0 & 95.0 & \textbf{100} & 98.0 \\
CDO & 87.0 & 99.9 & 93.9 & 81.5 & 94.1 & 98.0 & 97.0 & 96.1 & 89.1 & 99.4 & 95.6 & \textbf{100} & 99.0 & 99.5 & \textbf{100} & \textbf{100} \\
UniAD & 99.8 & 96.1 & 87.1 & 99.8 & 99.3 & 93.7 & 88.5 & 94.1 & 95.2 & 99.7 & 97.6 & 99.2 & 98.3 & 96.5 & 99.8 & \textbf{100} \\
\midrule
\textbf{Ours} & \underline{\textbf{94.0}} & \underline{91.1} & \underline{\textbf{99.4}} & \underline{91.8} & 92.2 & \underline{\textbf{99.2}} & \underline{\textbf{98.1}} & \underline{\textbf{99.6}} & \underline{\textbf{98.5}} & \underline{99.6} & \underline{\textbf{98.7}} & \textbf{100} & \textbf{100} & 95.0 & \textbf{100} & \textbf{97.1} \\
\bottomrule
\end{tabular}}
\end{table*}

The defect localization results on MVTec-AD are presented in \cref{tab:mvtec_loc}. FPFNet achieves an average pixel-level AUROC of 96.93\%, outperforming UniAD in 7 out of 15 categories, with particularly strong results on object-type categories. The average localization performance also exceeds other state-of-the-art methods, demonstrating the effectiveness of the multi-layer feature fusion strategy in capturing fine-grained spatial information necessary for precise defect localization.

\begin{table*}[t]
\centering
\caption{Defect localization accuracy (pixel-level AUROC, \%) of different methods on MVTec-AD. \textbf{Bold} indicates the best result per column.}
\label{tab:mvtec_loc}
\resizebox{\textwidth}{!}{
\begin{tabular}{l|cccccccccc|ccccc|c}
\toprule
\multirow{2}{*}{Method} & \multicolumn{10}{c|}{Object} & \multicolumn{5}{c|}{Texture} & \multirow{2}{*}{Mean} \\
& Bottle & Cable & Caps. & Haze. & M\_N. & Pill & Screw & Tooth. & Trans. & Zip. & Carpet & Grid & Leath. & Tile & Wood & \\
\midrule
DeSTSeg & 85.5 & 80.2 & 85.6 & 81.2 & 79.2 & 74.3 & 70.2 & 87.1 & 72.9 & 75.1 & 95.6 & 96.1 & 96.5 & 93.0 & 80.2 & 83.5 \\
DSR & 87.6 & 82.2 & 84.3 & 87.3 & 91.1 & 78.3 & 88.3 & 93.7 & 64.3 & 85.5 & 95.1 & 96.0 & 88.5 & 85.6 & 82.5 & 86.0 \\
ReContrast & 91.7 & 89.9 & 95.4 & 94.8 & 92.1 & 93.6 & 77.9 & 89.4 & 92.9 & 86.5 & 84.9 & 71.2 & 84.7 & 76.9 & 79.7 & 86.8 \\
DRAEM & 73.3 & 69.1 & 67.0 & 95.4 & 71.7 & 92.8 & 95.3 & 96.4 & 65.0 & 96.9 & 98.1 & 99.0 & 96.4 & 92.3 & 94.8 & 86.9 \\
BGAD & \textbf{97.8} & 82.8 & 95.9 & 97.6 & 93.1 & 97.8 & 95.1 & 80.1 & 97.7 & 96.1 & 95.0 & 98.6 & 98.2 & 99.0 & \textbf{99.4} & 95.8 \\
RD4AD & 95.6 & 85.7 & 98.3 & 93.1 & 97.7 & 87.0 & 97.4 & 98.1 & 99.3 & 95.0 & 95.6 & 95.9 & 98.8 & 99.1 & 98.8 & 99.3 \\
CDO & 97.2 & 95.9 & 97.6 & 98.1 & 97.1 & 98.1 & 97.3 & 95.5 & 96.5 & 93.8 & 96.5 & 93.1 & 89.8 & 96.3 & 98.8 & 98.6 \\
RD++ & 97.7 & 94.0 & 96.3 & 98.6 & 97.8 & 98.3 & 97.9 & 98.2 & 95.0 & 96.6 & 96.2 & 93.2 & 98.3 & 93.3 & 96.6 & 96.8 \\
UniAD & 98.5 & 98.0 & 93.4 & 95.1 & 98.4 & 98.4 & 98.2 & 96.6 & 98.5 & 96.6 & 98.8 & 92.1 & 93.2 & 96.7 & 98.0 & 97.3 \\
\midrule
\textbf{Ours} & 97.8 & 96.5 & \textbf{97.9} & 95.5 & 95.8 & \textbf{99.0} & 98.2 & 97.5 & 98.2 & 97.1 & \textbf{98.4} & 91.8 & 92.7 & \textbf{98.6} & 98.3 & \textbf{96.9} \\
\bottomrule
\end{tabular}}
\end{table*}

\subsubsection{Results on VisA}

The defect detection and localization results on VisA are reported in \cref{tab:visa_det} and \cref{tab:visa_loc}, respectively. On this more challenging benchmark, FPFNet demonstrates pronounced advantages, achieving 91.08\% image-level AUROC for detection and 99.08\% pixel-level AUROC for localization. The detection accuracy surpasses UniAD in 11 out of 12 categories, while the localization accuracy exceeds UniAD across all 12 categories, confirming the consistent effectiveness of our approach across different data distributions and defect characteristics.

\begin{table*}[t]
\centering
\caption{Defect detection accuracy (image-level AUROC, \%) of different methods on VisA. \textbf{Bold} indicates the best result.}
\label{tab:visa_det}
\resizebox{\textwidth}{!}{
\begin{tabular}{l|cccccccccccc|c}
\toprule
Method & candle & caps. & cashew & chew. & fryum & mac.1 & mac.2 & pcb1 & pcb2 & pcb3 & pcb4 & pipe\_f. & Mean \\
\midrule
DeSTSeg & 74.6 & 65.8 & 63.5 & 58.3 & 62.3 & 64.8 & 60.8 & 66.2 & 67.9 & 62.5 & 73.7 & 59.7 & 65.0 \\
ReContrast & 87.8 & 82.0 & 53.0 & 67.9 & 63.7 & 56.3 & 68.8 & 84.6 & 52.7 & 75.9 & 95.8 & 97.9 & 99.8 \\
DRAEM & 79.3 & 85.4 & 63.5 & 95.7 & 87.3 & 56.4 & 59.4 & 84.1 & 72.4 & 82.0 & 85.5 & 91.1 & 78.5 \\
DSR & 77.1 & 82.0 & 91.0 & 76.0 & 75.5 & 51.6 & 89.6 & 78.9 & 98.2 & 93.4 & 82.8 & 89.6 & 91.1 \\
RD++ & 88.2 & 75.8 & 91.2 & 81.8 & 88.3 & 77.3 & 61.9 & 90.8 & 84.8 & 80.9 & 99.2 & 93.1 & 84.4 \\
BGAD & 89.7 & 76.0 & 90.6 & 99.4 & 92.8 & 81.2 & 70.5 & 91.0 & 76.5 & 77.6 & 90.8 & 95.3 & 86.0 \\
CDO & 89.8 & 82.3 & 91.8 & 83.7 & 91.6 & 78.9 & 62.9 & 92.5 & 89.6 & 86.3 & 99.4 & 92.6 & 86.8 \\
RD4AD & 89.2 & 82.1 & 86.3 & 86.5 & 91.7 & 89.5 & 92.6 & 99.3 & 90.7 & 93.4 & 96.1 & 84.0 & 97.9 \\
UniAD & 93.5 & 70.9 & 89.7 & 98.5 & 87.5 & 88.1 & 77.6 & 93.2 & 90.5 & 85.4 & 98.8 & 96.0 & 89.1 \\
\midrule
\textbf{Ours} & 72.2 & \textbf{88.7} & \textbf{99.5} & 90.9 & \textbf{90.2} & \textbf{81.0} & \textbf{89.4} & \textbf{96.0} & \textbf{95.8} & \textbf{95.5} & \textbf{93.8} & \textbf{99.5} & \textbf{91.0} \\
\bottomrule
\end{tabular}}
\end{table*}

\begin{table*}[t]
\centering
\caption{Defect localization accuracy (pixel-level AUROC, \%) of different methods on VisA. \textbf{Bold} indicates the best result.}
\label{tab:visa_loc}
\resizebox{\textwidth}{!}{
\begin{tabular}{l|cccccccccccc|c}
\toprule
Method & candle & caps. & cashew & chew. & fryum & mac.1 & mac.2 & pcb1 & pcb2 & pcb3 & pcb4 & pipe\_f. & Mean \\
\midrule
DSR & 72.6 & 72.4 & 60.3 & 74.1 & 81.8 & 62.1 & 71.5 & 92.2 & 86.9 & 91.5 & 77.1 & 77.0 & 76.6 \\
DRAEM & 98.9 & 96.9 & 97.5 & 89.4 & 63.2 & 88.3 & 78.2 & 52.8 & 58.5 & 86.8 & 90.4 & 85.0 & 82.2 \\
DeSTSeg & 95.9 & 94.7 & 85.2 & 91.2 & 95.5 & 93.3 & 89.2 & 92.7 & 96.4 & 78.1 & 74.1 & 90.4 & 99.1 \\
ReContrast & 94.7 & 87.6 & 96.5 & 82.0 & 98.1 & 91.4 & 79.3 & 99.0 & 97.2 & 97.5 & 98.3 & 98.1 & 93.3 \\
RD++ & 98.4 & 97.1 & 96.1 & 87.0 & 98.7 & 91.2 & 92.8 & 98.3 & 97.8 & 97.7 & 98.6 & 99.6 & 96.1 \\
BGAD & 99.0 & 97.8 & 93.9 & 99.6 & 94.7 & 99.2 & 97.8 & 97.5 & 94.3 & 94.2 & 91.0 & 98.3 & 96.4 \\
RD4AD & 98.9 & 86.4 & 95.4 & 97.5 & 99.3 & 98.9 & 97.5 & 98.1 & 98.6 & 98.5 & 99.2 & 97.3 & 98.7 \\
CDO & 98.7 & 98.0 & \textbf{99.4} & 96.5 & 98.4 & 94.0 & 90.3 & 98.8 & 98.5 & \textbf{99.6} & 97.5 & 98.7 & 99.1 \\
UniAD & 99.3 & 97.4 & \textbf{99.4} & 99.3 & 98.6 & 99.2 & 97.2 & 99.6 & 98.2 & 98.6 & 98.8 & 99.6 & 98.8 \\
\midrule
\textbf{Ours} & 98.1 & \textbf{98.6} & 99.5 & \textbf{99.5} & \textbf{99.5} & \textbf{99.4} & 97.9 & \textbf{99.7} & \textbf{98.7} & 98.9 & \textbf{99.1} & \textbf{99.7} & \textbf{99.0} \\
\bottomrule
\end{tabular}}
\end{table*}

The results across both datasets reveal that FPFNet yields more pronounced improvements on the VisA dataset compared to MVTec-AD. This observation can be attributed to the greater diversity and complexity of the VisA dataset, where the benefits of diversified feature perturbation and multi-layer feature fusion are more evident. The consistent improvements in both detection and localization metrics across different product categories validate the generalization capability of our approach.

\subsection{Ablation Study}

To systematically validate the contribution of each proposed component, we conduct comprehensive ablation studies on both MVTec-AD and VisA datasets.

\subsubsection{Effect of Feature Perturbation Pool}

We compare the performance of individual perturbation types against the full perturbation pool, with results presented in \cref{tab:ablation_perturb}. On MVTec-AD, using F-Noise or F-Drop individually results in moderate performance degradation compared to the baseline Gaussian Noise. However, when the feature perturbation pool is employed, the detection accuracy improves to 96.81\% while maintaining comparable localization performance. On VisA, the perturbation pool yields improvements in both detection (89.58\%) and localization (98.87\%) metrics. These results confirm that the stochastic combination of multiple perturbation types provides complementary benefits that exceed those of any single perturbation scheme.

\begin{table}[t]
\centering
\caption{Ablation study on different feature perturbation strategies (\%).}
\label{tab:ablation_perturb}
\begin{tabular}{l|cc|cc}
\toprule
\multirow{2}{*}{Perturbation} & \multicolumn{2}{c|}{MVTec-AD} & \multicolumn{2}{c}{VisA} \\
& Det. & Loc. & Det. & Loc. \\
\midrule
F-Noise & 95.69 & 96.67 & 81.71 & 97.12 \\
F-Drop & 96.01 & 96.69 & 81.04 & 96.82 \\
Gaussian Noise & 96.58 & 96.77 & 89.19 & 98.81 \\
\textbf{Perturbation Pool} & \textbf{96.81} & \textbf{96.76} & \textbf{89.58} & \textbf{98.87} \\
\bottomrule
\end{tabular}
\end{table}

\subsubsection{Effect of Feature Fusion}

To determine the optimal fusion configuration, we systematically evaluate different fusion placements, as reported in \cref{tab:ablation_fusion}. Applying fusion exclusively within the decoder module yields performance comparable to the baseline, while fusion solely within the encoder module results in some performance degradation. However, when fusion is applied simultaneously in both the encoder and decoder modules, a synergistic effect emerges: the detection accuracy reaches 97.10\% on MVTec-AD and 89.99\% on VisA, with localization accuracy of 96.92\% and 98.96\%, respectively. These results demonstrate that bidirectional fusion across the encoding and decoding pathways is essential for capturing the full spectrum of hierarchical feature interactions.

\begin{table}[t]
\centering
\caption{Ablation study on feature fusion configurations (\%).}
\label{tab:ablation_fusion}
\begin{tabular}{l|cc|cc}
\toprule
\multirow{2}{*}{Fusion Strategy} & \multicolumn{2}{c|}{MVTec-AD} & \multicolumn{2}{c}{VisA} \\
& Det. & Loc. & Det. & Loc. \\
\midrule
No Fusion & 96.58 & 96.77 & 89.19 & 98.81 \\
Encoder Only & 96.00 & 96.49 & 83.65 & 98.32 \\
Decoder Only & 96.54 & 96.74 & 89.10 & 98.83 \\
\textbf{Encoder + Decoder} & \textbf{97.10} & \textbf{96.92} & \textbf{89.99} & \textbf{98.96} \\
\bottomrule
\end{tabular}
\end{table}

\subsubsection{Combined Analysis}

\cref{tab:ablation_combined} presents the results of combining different improvement strategies. Both the feature perturbation pool and multi-layer feature fusion independently contribute to performance gains. When combined, they yield the best overall performance: 97.17\% detection and 96.93\% localization on MVTec-AD, and 91.08\% detection and 99.08\% localization on VisA. Importantly, all variants maintain identical learnable parameter counts (7.17\,M), model sizes (29.43\,MB), and computational complexity (20.31\,GFLOPS), confirming that the proposed improvements introduce no additional computational overhead.

\begin{table}[t]
\centering
\caption{Ablation study combining different improvements (\%).}
\label{tab:ablation_combined}
\resizebox{\columnwidth}{!}{
\begin{tabular}{cc|cc|cc|ccc}
\toprule
\multirow{2}{*}{Pool} & \multirow{2}{*}{Fusion} & \multicolumn{2}{c|}{MVTec-AD} & \multicolumn{2}{c|}{VisA} & Params & Size & FLOPs \\
& & Det. & Loc. & Det. & Loc. & (M) & (MB) & (G) \\
\midrule
& & 96.58 & 96.77 & 89.19 & 98.81 & 7.17 & 29.43 & 20.31 \\
\checkmark & & 96.81 & 96.76 & 89.58 & 98.87 & 7.17 & 29.43 & 20.31 \\
& \checkmark & 97.10 & 96.92 & 89.99 & 98.96 & 7.17 & 29.43 & 20.31 \\
\checkmark & \checkmark & \textbf{97.17} & \textbf{96.93} & \textbf{91.08} & \textbf{99.08} & 7.17 & 29.43 & 20.31 \\
\bottomrule
\end{tabular}}
\end{table}

We observe that while both improvements are effective individually, their combined benefit on detection accuracy is slightly less than additive (particularly when comparing ``Fusion only'' to ``Pool + Fusion''). This can be attributed to the ``learning shortcut'' phenomenon~\cite{you2022unified}: the model may partially reconstruct perturbed defect features as normal features, a tendency that is amplified when multi-layer fusion propagates these partially learned defect features from shallow to deep layers. Nevertheless, the consistent improvements across all metrics confirm the overall effectiveness of the proposed approach.

\subsection{Analysis and Discussion}

\noindent\textbf{Performance on Texture vs.\ Object Categories.} Our method exhibits particularly strong performance on texture-type categories in terms of detection accuracy, while demonstrating advantages on object-type categories for localization. The perturbation pool appears to be especially beneficial for texture categories, where the distinction between normal and anomalous patterns is often subtle. For object categories, the multi-layer feature fusion mechanism provides the primary benefit by enabling more precise spatial reasoning.

\noindent\textbf{Failure Case Analysis.} Despite the strong overall performance, we observe that for certain texture-type samples, the localization accuracy shows a slight decrease compared to UniAD. Analysis reveals that texture defect features often closely resemble normal texture features, and while the proposed perturbation pool and feature fusion enhance the model's generalization capability, they may also slightly blur the distinction between very similar normal and defective texture patterns in the localization task. This represents an inherent challenge in texture anomaly detection that warrants further investigation.

\noindent\textbf{Cross-Dataset Generalization.} The more pronounced improvements on VisA compared to MVTec-AD suggest that FPFNet's benefits scale with dataset complexity and diversity. The VisA dataset, with its larger variety of product types and more complex defect patterns, provides a more favorable setting for the diversified perturbation and hierarchical fusion strategies to demonstrate their advantages. This finding highlights the potential of our approach for real-world deployment scenarios involving diverse product portfolios.

\section{Conclusion}
\label{sec:conclusion}

In this paper, we have presented FPFNet, a Feature Perturbation Pool-based Fusion Network for unified multi-class industrial defect detection that addresses two critical limitations of existing approaches through two synergistic innovations: a feature perturbation pool that stochastically applies diverse noise patterns (Gaussian Noise, F-Noise, and F-Drop) to enrich the training distribution and enhance model robustness, and a multi-layer feature fusion strategy that aggregates hierarchical feature representations from both encoder and decoder modules through residual connections and normalization to capture complex cross-scale feature relationships. Built upon the UniAD architecture, FPFNet achieves state-of-the-art performance on MVTec-AD (97.17\% detection AUROC, 96.93\% localization AUROC) and VisA (91.08\% detection AUROC, 99.08\% localization AUROC) without introducing any additional learnable parameters or computational overhead, demonstrating that principled perturbation diversification and lightweight hierarchical fusion can yield substantial performance gains within a parameter-free improvement paradigm. Future work will explore more sophisticated feature enhancement techniques, attention-based mechanisms for defect feature distribution modeling, and the extension of the proposed framework to handle an even broader range of defect types and more complex production environments, ultimately contributing to the evolution of industrial quality control systems toward smarter and more efficient operation.

{
\small
\noindent\textbf{Acknowledgments.}
This work was supported by the National Natural Science Foundation of China (Grant No.\ U20A20228).
}

{\small
\bibliographystyle{ieee_fullname}
\bibliography{references}
}

\end{document}